\documentclass{article}

\PassOptionsToPackage{numbers, compress}{natbib}

\usepackage{natbib}
\usepackage[preprint]{neurips_2025}

\usepackage[utf8]{inputenc} 
\usepackage[T1]{fontenc}    
\usepackage{hyperref}       
\usepackage{url}            
\usepackage{booktabs}       
\usepackage{amsmath}
\usepackage{amsfonts}       
\usepackage{nicefrac}       
\usepackage{microtype}      
\usepackage{xcolor}         

\usepackage{epigraph}  
\usepackage{lipsum}    
\usepackage{graphicx}  
\usepackage{geometry}  
\usepackage{caption}   
\usepackage{subcaption}
\usepackage[most]{tcolorbox}
\usepackage{wrapfig}
\usepackage{multirow}
\usepackage{float}
\usepackage{algorithm2e}
\usepackage{listings}

\usepackage[normalem]{ulem}

\title{\method: \textbf{Ca}usal \textbf{R}easoning \textbf{G}raph-\textbf{o}f-\textbf{T}hought improves Multimodal Humor Comprehension}

\author{%
Abhilash Nandy\thanks{This work was carried out while the author was affiliated with Indian Institute of Technology Kharagpur, India} \\
Microsoft Research India \\
\And
Rahul Seetharaman \\
LinkedIn, USA \\
\And
Aman Bansal \\
Nutanix, USA \\
\AND
Rounak Saha \\
Indian Institute of Science, India \\
\And
Manav Nitin Kapadnis \\
Apple, USA \\
\And
Millon Madhur Das \\
Fujitsu Research India \\
\AND
Pawan Goyal, Niloy Ganguly \\
Indian Institute of Technology Kharagpur, India \\
}

\begin{document}

\newcommand{\noteng}[1]{}
\newcommand{\pg}[1]{}
\newcommand{\abhi}[1]{}
\newcommand{\rounak}[1]{}

\newcommand{\method}{\textsc{CaRGo-T}}

\definecolor{hlred}{RGB}{255,200,200}
\definecolor{hlgreen}{RGB}{200,255,200}
\definecolor{darkred}{RGB}{139,0,0}    
\definecolor{darkgreen}{RGB}{0,100,0}  

\DeclareRobustCommand{\hlr}[1]{{\sethlcolor{hlred}\hl{#1}}}
\DeclareRobustCommand{\hlg}[1]{{\sethlcolor{hlgreen}\hl{#1}}}

\newcommand{\Ruline}[1]{\textcolor{red}{\uline{#1}}}
\newcommand{\Guline}[1]{\textcolor{green!60!black}{\uline{#1}}}

\lstdefinestyle{jsonstyle}{
  basicstyle=\ttfamily\scriptsize,
  numbers=none,
  showstringspaces=false,
  breaklines=true,
  moredelim=**[is][\colorbox{hlred}]{@R}{@},
  moredelim=**[is][\colorbox{hlgreen}]{@G}{@},
  moredelim=**[is][\colorbox{darkred}]{@S}{@},
  moredelim=**[is][\colorbox{darkgreen}]{@H}{@}  
}

\maketitle
\setcounter{footnote}{0}

\begin{abstract}

In recent years, large-scale VLMs (Vision-Language Models) have achieved exceptional versatility and performance across a diverse array of tasks. However, these systems continue to face significant challenges in capturing and interpreting the subtle complexities of human humor, especially in a combination of both image and text modalities \pg{Are you focusing specifically on combination of text and image? Is this problem easier when only one modality is invilved?}, as they involve complex, non-trivial interactions between people, objects, abstract concepts, and events, which constitute the foundational basis for a multitude of comedic mechanisms and humor signals. Causal graphs provide a natural formalism for representing the intertwined chains of events, entities, and contextual signals that give rise to humor in multimodal content. In this paper, we propose \method\ (\textbf{Ca}usal \textbf{R}easoning \textbf{G}raph-\textbf{o}f-\textbf{T}hought), which uses a graph serialized into a VLM-generated lightweight, code‐based reasoning component that the same/different VLM can interpret to provide the final answer in a zero-shot/in-context learning setting. Through rigorous experimentation of \method\ on state‑of‑the‑art commercial and open‑source large VLMs on humor understanding and detection tasks—employing four datasets that span diverse comedic dimensions such as satire, sarcasm, and memes. \method\ improves performance by $\sim1-20\%$ in Humor Understanding and $\sim1-3\%$ in Humor Detection compared to other reasoning-based baselines. Further analysis on mutual information of the generated reasoning component reveals that \method\ offers more information compared to baselines that is relevant to the desired output\footnote{Code for the work is available at \url{https://github.com/abhi1nandy2/CaRGo-T}}.

\end{abstract}

\section{Introduction}
\label{sec:intro}

\begin{minipage}[t]{\linewidth}
\renewcommand{\epigraphflush}{flushleft}
\setlength{\epigraphwidth}{\linewidth}
\epigraph{\itshape ``Analyzing humor is like dissecting a frog. Few people are interested and the frog dies of it.''}{---E. B. White}
\end{minipage}

Humor intricately weaves together the relationships among people, objects, concepts, and events, revealing subtle incongruities that both amuse and illuminate our understanding of everyday interactions. Effective humor comprehension similarly relies on advanced social reasoning and cultural literacy, as jokes often draw upon shared norms, stereotypes, and contextual cues to create meaning. Additionally, humor can employ nonlinear narrative structures and symbolic incongruities—much like comics—demanding sophisticated reasoning to detect and appreciate the unexpected connections that make a joke resonate \cite{pressman2014digital,manning1998understanding}. Contemporary large-scale VLMs have demonstrated exceptional capabilities across a wide range of applications \cite{liu2023improved,yin2023survey,wu2023multimodal}. However, prior works which measure the ability of SOTA VLMs to understand humor in multimodal settings show that such VLMs fail at comprehending different kinds of humor such as satire, irony, sarcasm, etc. satisfactorily \cite{nandy-etal-2024-yesbut, qin2023mmsd2, hwang-shwartz-2023-memecap}.


\pg{There is some disconnect.} VLMs demonstrate significant shortcomings when humor derives from intricate social relations and object-mediated interactions. Such scenarios demand fine-grained modeling of affective cues and relational incongruities, which current architectures fail to represent adequately. For instance, in collaborative scenes requiring subtle appraisal of interpersonal dynamics and object-mediated interactions, current VLMs exhibit pronounced failures: they misclassify core emotional states (e.g., mistaking sarcasm for surprise) in dyadic exchanges and group settings \cite{bhattacharyya2025evaluating}, overlook incongruous action–object relationships that give rise to visual punchlines \cite{hu2024cracking}, and collapse multimodal figurative content into single literal interpretations, thereby missing alternate relational perspectives essential for nuanced humor \cite{saakyan2024understanding}. Moreover, explicit reasoning augmentations fail to bridge these gaps: Chain‐of‐Thought \cite{cot} prompting suffers posterior collapse in subjective, emotion‐laden contexts—retrieving static priors rather than dynamically reasoning over interpersonal cues—and yields negligible improvement on sarcasm and affect recognition benchmarks \cite{chochlakis2024larger}. Likewise, multimodal extensions such as self‐reflection frameworks \cite{deng2024enhancing} produce noisy, misaligned rationales that overlook the fine‐grained affective and relational subtleties at the heart of social humor comprehension. Hence, there is a need to understand complex relations in scenarios involving interplay of multiple people, objects, concepts and/or events in a multimodal setting.

\begin{wrapfigure}{tr}{0.62\textwidth}
    \centering
    \includegraphics[width=\linewidth]{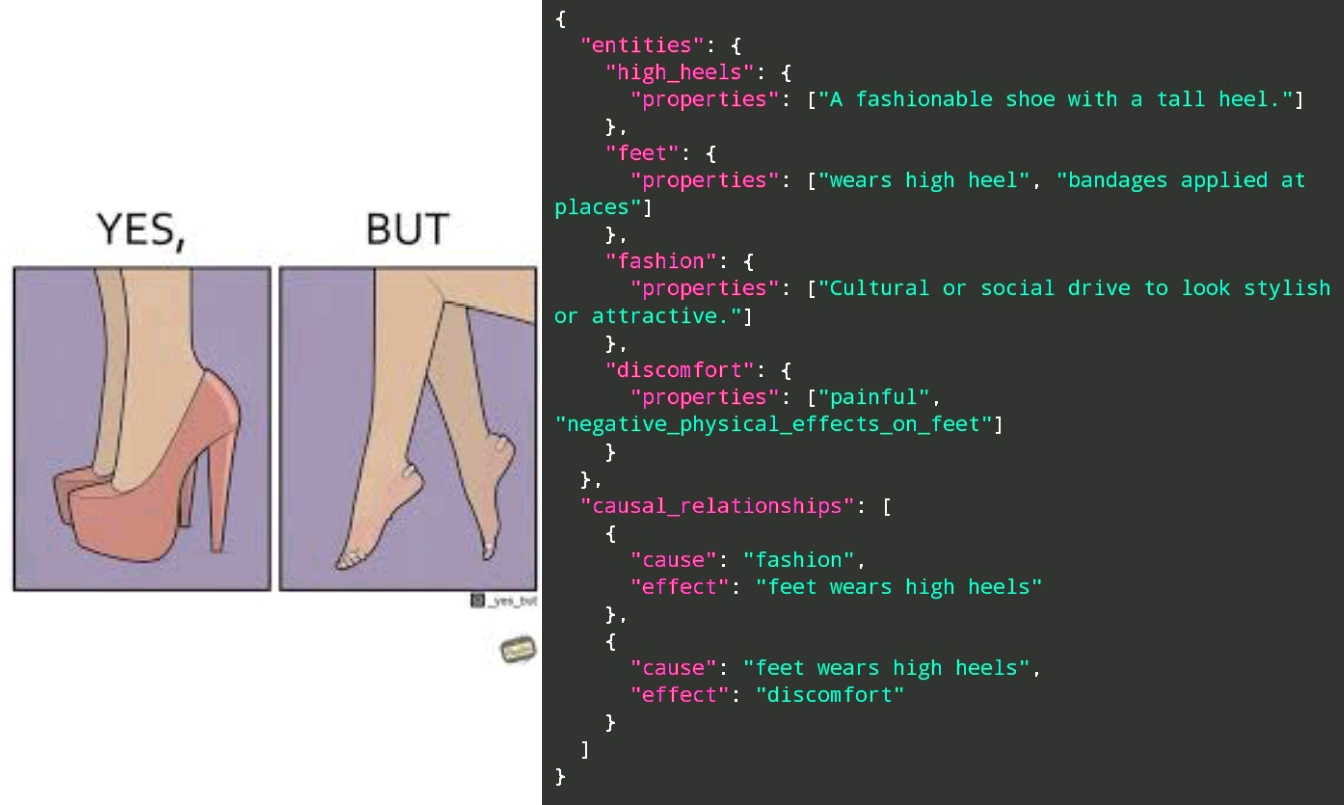}
    \caption{Sample Causal Reasoning Graph for an image from the \textit{\textbf{YesBut}} Dataset \cite{nandy-etal-2024-yesbut}. The image shows that in order to look fashionable, a person is wearing high heels, which in turn leads to discomfort. This is depicted using the cause-effect relations in the Causal Reasoning Graph, in addition to the description of objects (high heels, feet) and abstract concepts (fashion, discomfort) in the given image} 
    \label{fig:intro_example}
\end{wrapfigure} 

In order to understand and navigate through those intricacies, in this paper, we introduce {\em causal reasoning graphs}, a restricted subclass of causal graphs \cite{helmert2004planning} that retains only cause–effect links—together with lightweight metadata describing objects, concepts, events, and participants—without any probabilistic parameters. These deterministic, event‐centric structures serve as an interpretable backbone for modeling multifaceted multimodal scenarios. To the best of our knowledge, no prior work in multimodal humor comprehension has explicitly modeled causal graphs or cause–effect relationships to enhance humor understanding; we are the first to undertake this effort. (See Fig. \ref{fig:intro_example} for an example of a causal reasoning graph)

In order to automatically construct these CRGs and subsequently use it for humor detection and understanding task, 
we propose \method, which uses the power of VLM to first construct an explicit causal reasoning graph encoding the interplay of agents, objects, concepts, and events (along with relevant metadata) in the form of a lightweight, code-based reasoning script (Fig. \ref{fig:intro_example} for reference). 
The causal reasoning graph is then ingested by the same/different VLM in a zero-shot/in-context-learning setting to detect and understand humor. 
The novelty of \method\ lies in using a causal reasoning graph as the reasoning component, instead of using natural language reasoning in Chain-of-Thought and Chain-of-Draft \cite{cot, cod}, or using knowledge graph triplets in \citet{kgreasoning}. The causal reasoning graph  enables systematic causal traversal and compositional inference that accurately captures the dynamics and relational incongruities, which standard chain-of-thought \cite{cot, zeroshotcot} and other reasoning-based baselines \cite{cod, mitra2024compositional, kgreasoning} fail to capture.

Extensive evaluations on three benchmark datasets—spanning satire, sarcasm, and visual memes —  demonstrate that \method\ consistently outperforms contemporary reasoning-based baselines by a significant margin of $\sim1-20\%$ in Humor Understanding and $\sim1-3\%$ in Humor Detection across different VLMs and settings. Furthermore, an information-theoretic analysis reveals that the generated reasoning component under \method\ contains more information than the baselines, and is also more relevant to the ground truth, offering stronger task-relevant supervision. Our key contributions are as follows: (i) we propose \method, a novel VLM-agnostic framework for constructing and leveraging explicit causal reasoning scripts in visual humor tasks; (ii) we show that \method\ enables systematic causal traversal and compositional inference beyond standard CoT and baseline methods; (iii) we conduct comprehensive experiments on four diverse humor datasets, achieving substantial performance gains; and (iv) we provide an in-depth mutual information analysis that quantifies the superiority of \method\ compared to baselines, and relevance of our reasoning component to the target outcome.

\section{Related Work}

\noindent \textbf{LLMs and VLMs. }Large-scale language and vision models have shown exceptional ability to follow human instructions and address a wide array of downstream tasks through zero-shot prompting~\cite{ouyang2022training,llama3modelcard,minaee2024large,yin2023survey}. To systematically evaluate these advances, a variety of benchmarks have been developed, covering purely linguistic challenges~\cite{zheng2023judging,dubois2024alpacafarm,wang2024pandalm,huang2024c} as well as multimodal evaluations tailored for VLMs~\cite{ying2024mmt,bitton2023visit,bitton2023breaking,li2024seed,li2023seeda}. However, such models are still inadequate at understanding intricate social reasoning and human contexts \cite{nandy-etal-2024-yesbut, bhattacharyya2025evaluating, saakyan2024understanding, chochlakis2024larger}.

\noindent \textbf{Humor Comprehension and AI. }Humor serves as a fundamental aspect of human interaction~\cite{palmer2003taking}, prompting extensive research into tasks like humor recognition~\cite{chen2017predicting,cattle2018recognizing,yang2015humor} and humor generation~\cite{amin-burghardt-2020-survey}. Building on these foundations, recent studies have ventured into multimodal domains—forecasting visual humor~\cite{chandrasekaran2016we}, pinpointing humorous cartoon/meme captions~\cite{hwang-shwartz-2023-memecap,hessel-etal-2023-androids,radev2016humor}, and detecting humor in video content~\cite{nandy-etal-2024-yesbut,liu2024comment}. Nonetheless, despite these strides, LLMs such as ChatGPT are yet to fully conquer the complexities of computational humor~\cite{jentzsch2023chatgpt}.

\noindent
\textbf{Reasoning on Vision and Language.} We focus on tasks related to humor comprehension in multimodal scenarios, where the model requires in-depth reasoning to comprehend the interactions between different parts of the image and text in the inputs. Prior Art has benchmarked reasoning capabilities of state-of-the-art models in tasks related to commonsense reasoning \cite{bitton2023breaking}, visual question answering~\cite{hudson2019gqa}, and visio-linguistic compositionality~\cite{thrush2022winoground}.

\noindent \textbf{Causal Reasoning and AI.} A recent comprehensive survey \cite{liu2403large} recasts the evaluation of large language models within a causal framework, examining their inferential strengths, strategies for mitigating bias and ensuring safety, methods for enriching outputs with interpretable explanations, and extensions into multimodal domains. Building on this causal perspective, \citet{chen2024clear} introduces a rigorous framework for assessing causal graph understanding in language models, grounded in four practical criteria drawn from philosophy and psychology. They further propose CLEAR, a benchmark on which LLMs exhibit early signs of causal comprehension, along with considerable room for improvement. In CausE \cite{zhang2023cause}, the focus pivots from traditional NLP tasks to the knowledge graph completion task, employing causal intervention and embedding disentanglement to mitigate confounders and achieve more stable predictions. CELLO \cite{chen2024cello} is a multimodal benchmark that evaluates VLMs on several multiple-choice causal questions. However, no prior art to our knowledge has utilized Cause-Effect Relations for boosting open-ended multimodal reasoning, which is the focus of this paper.

\section{\method\ Framework}
\label{sec:generic-reasoning}

Given an image $\mathcal{I}$ and a task-specific text prompt $\mathcal{P}$, the predicted answer is $\mathcal{\hat{Y}} \subseteq \mathcal{F}_{\boldsymbol{\theta}}(\mathcal{I}, \mathcal{P})$. The VLM output $\mathcal{F}_{\boldsymbol{\theta}}(\mathcal{I}, \mathcal{P})$ contains a reasoning component $\mathcal{R}$, followed by $\mathcal{\hat{Y}}$, where $\mathcal{F}_{\boldsymbol{\theta}}$ represents the pre-trained VLM with parameters ${\boldsymbol{\theta}}$. This is a generalized representation of any reasoning-based method such as CoT, CoD etc. 
As a replacement we introduce {\bf \method\ }  a novel code-based reasoning framework using pre-trained VLMs on multimodal (vision-cum-image) inputs in a zero-shot/in-context learning setting to perform tasks related to humor comprehension. 


\subsection{\method: Zero-Shot Setting}

The input prompt for zero-shot setting is given below for \method\ (task-specific placeholder is written in \textcolor{red}{red}). The prompt starts with the vanilla zero-shot task-specific query from prior art (e.g. For the Satirical Image Understanding Task evaluated in \citet{nandy-etal-2024-yesbut}, the vanilla zero-shot task-specific query used is - ``Why is this image funny/satirical?''; all task-specific queries are listed \textit{\textbf{in Section \ref{sec:task-spec-q} of Appendix}}). We then extend this base query by requesting the model to first construct a causal reasoning graph, represented in code, that captures the relationships among objects, people, and entities depicted in the image (and input text, if any, depending on the task), and subsequently produce the final answer for the task conditioned on the underlying cause-and-effect structure. By structuring the prompt into two distinct steps—graph generation followed by interpretation—we encourage the model to make its latent causal reasoning explicit before formulating the final answer. 

\begin{tcolorbox}[
  enhanced,
  colback=gray!20,      
  colframe=black,       
  width=\textwidth,     
  boxrule=0.5pt,        
  arc=0pt,              
  outer arc=0pt,
  title={\color{white}\method\ Zero-Shot Prompt}, 
  fonttitle=\bfseries,  
  left=1em,             
  right=1em,
  top=1em,
  bottom=1em
]
{\small
\textcolor{red}{[VANILLA TASK-SPECIFIC QUERY]}

To answer this, first create a causal reasoning graph linking different objects, people, and entities present in the image (and input text, if any) in the form of a piece of code, and then give the final answer.

Input Image: \textcolor{red}{[INPUT IMAGE]}

Input Text: \textcolor{red}{[INPUT TEXT]}

The output should be in the following format -

Code: <Causal Reasoning Graph Code>

Final Answer: <Final Answer>
}
\end{tcolorbox}

Note that using \method\ in a zero-shot setting depends on the code-generation capability of the VLM, leveraging the VLM's parametric knowledge. Based on the evaluation of code generated using VLMs in prior work \cite{codevision}, proprietary models such as GPT-4o and GPT-4o-mini \cite{gpt-4o} are far better at code generation in a zero-shot setting compared to open-source models. Hence, we mostly use closed-source proprietary VLMs for the \method\ framework in the zero-shot setting.

\subsection{\method: In-Context Learning}

\method\ in an in-context learning setting contains input-output pairs as in-context examples in the prompt, where the input consists of the input image and text (if any), and the output consists of the corresponding Causal Reasoning Graph followed by the answer. This is described in more detail as follows -

\subsubsection{Curating In-Context Examples}
\label{sec:curate-ice}

Some dataset samples that are not part of the test set (e.g. from the training set) are used for curating the in-context examples. The input image (and text if any) and the ground truth final answer are available directly in the dataset sample. The corresponding Causal Reasoning Graph is curated by first generating a draft using the following zero-shot prompt (task and sample-specific placeholders are written in \textcolor{red}{red}), which generates said graph conditioned on the input and the final answer using GPT-4o. 

\begin{tcolorbox}[
  enhanced,
  colback=gray!20,      
  colframe=black,       
  width=\textwidth,     
  boxrule=0.5pt,        
  arc=0pt,              
  outer arc=0pt,
  title={\color{white} Prompt for generating an initial draft of a Causal Reasoning Graph for an in-context example}, 
  fonttitle=\bfseries,  
  left=1em,             
  right=1em,
  top=1em,
  bottom=1em
]
{\small
\textcolor{red}{[VANILLA TASK-SPECIFIC QUERY]}

To answer this, first create a causal reasoning graph linking different objects, people, and entities present in the image (and input text, if any) in the form of a piece of code, and then give the final answer. 

Input Image: \textcolor{red}{[INPUT IMAGE]}

Input Text: \textcolor{red}{[INPUT TEXT]}

The output is - 

Code: <Causal Reasoning Graph Code> 

Final Answer: \textcolor{red}{[GROUND TRUTH FINAL ANSWER]}}
\end{tcolorbox}

These graphs are then rectified manually to adhere to a specific structure - starting by identifying the entities (an entity could either be an object, person, abstract concept, or event), listing the properties of each entity, and finally listing the cause-effect relations, where each relation is a cause-effect pair derived from entities and their corresponding properties (see Fig. \ref{fig:intro_example}). An example of this manual rectification of causal reasoning graphs is shown \textit{\textbf{in Fig. \ref{fig:rectify-crg} in Section \ref{sec:manual-rectify} of Appendix}}.

\subsubsection{In-Context Learning Setup}

\begin{tcolorbox}[
  enhanced,
  colback=gray!20,      
  colframe=black,       
  width=\textwidth,     
  boxrule=0.5pt,        
  arc=0pt,              
  outer arc=0pt,
  title={\color{white}\method: K-Shot In-Context Learning Prompt}, 
  fonttitle=\bfseries,  
  left=1em,             
  right=1em,
  top=1em,
  bottom=1em
]
{\small
\textcolor{red}{[VANILLA TASK-SPECIFIC QUERY]}

To answer this, first create a causal reasoning graph linking different objects, people, and entities present in the image (and input text, if any) in the form of a piece of code, and then give the final answer.

\#\# Example 1

Input Image: \textcolor{red}{[INPUT IMAGE \#1]}

Input Text: \textcolor{red}{[INPUT TEXT \#1]}

The output is -

Code: \textcolor{red}{[CAUSAL REASONING GRAPH CODE \#1]}

Final Answer: \textcolor{red}{[GROUND TRUTH FINAL ANSWER \#1]}

..

..

\#\# Example \#\textcolor{red}{K}

// Exactly same format as Example \#1





\#\# Example \#\textcolor{red}{K+1}

Input Image: \textcolor{red}{[INPUT TEST IMAGE]}

Input Text: \textcolor{red}{[INPUT TEST TEXT]}

The output should be in the following format -

Code: <Causal Reasoning Graph Code>

Final Answer: <Final Answer>
}
\end{tcolorbox}

For a test input, K in-context examples (in a K-shot setting) consisting of input images (and texts, if any), manually rectified causal graphs and the ground truth final answers and task-specific instructions are added to the prompt as shown above. With this prompt as input, The VLM is expected to generate a causal graph and the final answer corresponding to the test input at hand. Note that the VLM is not trained; it is used only for inference.

\section{Experiments and Results}
\label{sec:expts-and-results}

In this section, we first introduce the tasks and datasets used in our study, then describe the experimental setup and baseline methods, outline the evaluation metrics, present comparisons with those baselines, and conclude with an in-depth examination of \method’s reasoning component. In addition, we also perform an ablation study \textit{\textbf{in Section \ref{sec:ablation} of Appendix}} to show the importance of task-specific query and manually rectifying in-context examples in \method.

\subsection{Tasks and Datasets}

We focus on tasks related to Humor Understanding (answering why the input is funny) and Humor Detection (answering whether the input is funny or not) across different types of humor. Any such task can be formally written as in Section \ref{sec:generic-reasoning} - $\mathcal{R}$ is empty if there is no reasoning component. The ground truth answer $\mathcal{Y}$   for Humor Understanding is in natural language, while for Humor Detection, $\mathcal{Y} \in \{\text{``Yes''}, \text{``No''}\}$  and 
and  the output $\mathcal{\hat{Y}}$  should ideally be highly similar to $\mathcal{Y}$.
Examples of such tasks are shown \textit{\textbf{in Fig. \ref{fig:humor-example} of Section \ref{sec:humor-example} in Appendix}}.

Tasks in Humor Understanding include - (1) \textit{Satirical Image Understanding: }We evaluate a VLM's satire understanding capability on satirical images of the \textit{\textbf{YesBut}} Dataset \cite{nandy-etal-2024-yesbut} by prompting the VLMs to generate the corresponding punchlines. A holdout set of $5$ diverse examples is manually selected for in-context example selection, and the rest of the $1,079$ satirical images are used for evaluation, (2) \textit{Meme Caption Generation: }Given a meme image and title as input, the VLM needs to predict why that meme is funny. This is evaluated on $559$ test samples from the \textit{\textbf{MemeCap}} Dataset \cite{hwang-shwartz-2023-memecap}, and the in-context examples for in-context learning are manually selected from the corresponding training set. 

Similarly, tasks in Humor Detection include - (1) \textit{Satirical Image Detection: }This is a binary classification task in which the VLM must determine whether a given image is satirical or not. Similar to the Satirical Image Understanding Task, a holdout set of $6$ ($3$ satirical, $3$ non-satirical) samples is manually chosen for in-context example selection, and evaluation is done on $2,541$ ($1,081$ satirical and $1,460$ non-satirical) images of the \textit{\textbf{YesBut}} Dataset \cite{nandy-etal-2024-yesbut} (2) \textit{Multimodal Sarcasm Detection: }This is also a binary classification task in which the VLM must determine whether a given image supplemented with a supporting text is sarcastic or not. This is evaluated on $2,409$ ($1,037$ sarcastic and $1,372$ non-sarcastic) test samples of the \textit{\textbf{MMSD 2.0}} Dataset \cite{qin2023mmsd2}. In-context examples for in-context learning are manually selected from the corresponding training set. 

\subsection{Experimental Setup}

\method\ and the baselines are evaluated using the proprietary VLMs of GPT-4o and GPT-4o-mini \cite{gpt-4o} and the open-source MiniCPM\footnote{\url{https://huggingface.co/openbmb/MiniCPM-V-2_6}} VLM \cite{yao2024minicpm}. The choice behind using these VLMs is inspired by the VLMs that are good at code generation as shown when evaluated on the Code-Vision Benchmark \cite{codevision}. All experiments using open-source models are performed on 2 NVIDIA L40 GPUs, each having a VRAM of 48GB. 

\subsection{Baselines}

\method\ is compared to the following VLM-agnostic baseline frameworks - (1) \textbf{Vanilla. }This baseline uses a prompt consisting of task-specific query and instructions. This is carried out in both zero-shot and few-shot in-context learning settings. In-Context Examples are added in the few-shot setting prompt (2) \textbf{CoT (Chain-of-Thought) \cite{cot, zeroshotcot} -} CoT prompting asks the model to articulate intermediate reasoning steps before arriving at an answer, effectively turning black-box inference into an explicit, human-readable chain of deductions. Note that in the in-context learning setting, the rationale (reasoning component of CoT) of an in-context example is generated in a way similar to that in Section \ref{sec:curate-ice}, except that the rationale is unstructured (see Section \ref{appendix:cot} for an example)
(3) \textbf{CoD (Chain-of-Draft) \cite{cod} -} CoD is a prompting paradigm in which a model produces concise intermediate ``draft'' summaries of its reasoning—retaining only indispensable information—before arriving at a final answer, thereby reducing verbosity while preserving the logical structure. It is applied in a zero-shot setting 
(4) \textbf{CCoT (Compositional Chain-of-Thought) \cite{mitra2024compositional} -} CCoT is a zero-shot prompting technique that leverages automatically generated scene-graph representations to guide large multimodal models in capturing object attributes and inter-object relationships. By injecting the inferred scene graph into the prompt, it elicits structured, compositional reasoning without  additional fine-tuning. 

\subsection{Evaluation Metrics}

For the Humor Understanding Tasks, we compare the VLM generated text with the ground truth using automated text comparison metrics - lexical overlap using BLEU \cite{bleu} and ROUGE-L \cite{lin-2004-rouge}, and semantic similarity using BERTScore \cite{bertscore}, along with an ``Avg. Score'' that is the mean of these $3$ metrics \footnote{Note that for Satirical Understanding on \textit{\textbf{YesBut}} Dataset \cite{nandy-etal-2024-yesbut} the individual metrics are averaged across the $3$ stages (each stage has a different combination of image styles) of the Dataset}. For the Humor Detection Tasks, we use binary classification evaluation metrics of Accuracy and macro-F1 Score.

\subsection{\method\ on Humor Understanding}

\subsubsection{Zero-shot Setting}

\begin{wrapfigure}{tr}{0.5\textwidth}
    \centering
    \includegraphics[width=\linewidth]{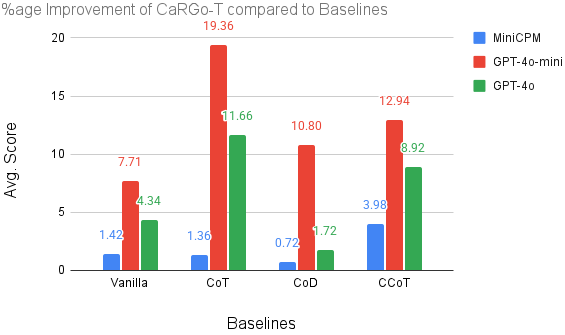}
    \caption{Percentage Improvement of \method\ compared to Baselines in zero-shot setting when evaluating on Satirical Image Understanding}
    \label{fig:yesbut_impr}
\end{wrapfigure}

Table \ref{tab:humor-und} shows the performance of \method\ vs. several baselines on the Satirical Image Understanding and Meme captioning Tasks in zero-shot setting across different VLMs. The corresponding percentage improvements of \method\ in comparison to the baselines are displayed in Fig. \ref{fig:yesbut_impr}. 

\begin{table}[H]
\centering
\resizebox{\linewidth}{!}{%
\begin{tabular}{c|c|cccc|cccc}
\hline
\multirow{2}{*}{\textbf{VLM}} & \multirow{2}{*}{\textbf{Method}} & \multicolumn{4}{c|}{\textbf{Satirical Image Understanding}} & \multicolumn{4}{c}{\textbf{Meme Captioning}} \\ \cline{3-10}
                             &                                  & \textbf{ROUGE-L} & \textbf{BLEU} & \textbf{BERTScore} & \textbf{Avg.} & \textbf{ROUGE-L} & \textbf{BLEU} & \textbf{BERTScore} & \textbf{Avg.} \\ \hline 

\multirow{6}{*}{\begin{tabular}[c]{@{}c@{}}MiniCPM\\(0-shot)\end{tabular}} 
& Vanilla    & 0.1669 & 0.0108 & 0.8589 & 0.3455 & 0.0789 & \textit{0.0073} & 0.833 & 0.3064 \\
& CoT        & 0.163 & \textbf{0.0155} & 0.8586 & 0.3457 & 0.0646 & 0.0033 & 0.8303 & 0.2994 \\
& CoD          & \textit{0.1684} & 0.0137 & \textbf{0.8616} & \textit{0.3479} & \textit{0.0888} & 0.0059  & 0.8394 & \textit{0.3114} \\ 
& CCoT       & 0.1482 & 0.0086 & 0.8541 & 0.337 & 0.0744 & 0.0040 & \textit{0.8404} & 0.3063 \\
& \method       & \textbf{0.1779} & \textit{0.0139} & \textit{0.8594} & \textbf{0.3504} & \textbf{0.126} & \textbf{0.0123} & \textbf{0.8503} & \textbf{0.3295} \\ \hline \\ \hline

\multirow{6}{*}{\begin{tabular}[c]{@{}c@{}}GPT-4o-mini\\(0-shot)\end{tabular}} 
& Vanilla    & \textit{0.1493} & 0.0098 & \textit{0.8525} & \textit{0.3372} & 0.0889 & 0.0045 & 0.845 & 0.3128 \\
& CoT        & 0.088 & 0.0052 & 0.8197 & 0.3043 & 0.1178 & 0.0076 & 0.8453 & 0.3236   \\
& CoD          & 0.1388 & \textit{0.0107} & 0.834 & 0.3278 & 0.1282 & \textit{0.0078} & 0.8459 & 0.3273    \\
& CCoT       & 0.1184 & 0.0062 & 0.8402 & 0.3216 & \textit{0.1315} & 0.0069 & \textbf{0.8566} & \textit{0.3317}   \\
& \method       & \textbf{0.2024} & \textbf{0.0185} & \textbf{0.8687} & \textbf{0.3632} & \textbf{0.1377} & \textbf{0.008} & \textit{0.8497} & \textbf{0.3318} \\ \hline \\ \hline

\multirow{6}{*}{\begin{tabular}[c]{@{}c@{}}GPT-4o\\(0-shot)\end{tabular}} 
& Vanilla    & 0.1893 & 0.0152 &  0.8667 & 0.3571 & 0.1092 & 0.006 & 0.8512 & 0.3221 \\
& CoT        & 0.1459 & 0.0082 & 0.847 & 0.3337 & 0.1183 & 0.0064 & 0.8523 & 0.3257   \\
& CoD          & \textit{0.2064} & \textit{0.02} & \textbf{0.8725} & \textit{0.3663} & \textit{0.1264} & \textit{0.0069} & \textit{0.8535} & \textit{0.3289}    \\
& CCoT       & 0.1605 & 0.0094 & 0.8564 & 0.3421 & 0.1055 & 0.0057 & \textit{0.8535} & 0.3216   \\
& \method       & \textbf{0.2219} & \textbf{0.0245} & \textit{0.8715} & \textbf{0.3726} & \textbf{0.1316} & \textbf{0.0078} & \textbf{0.8569} & \textbf{0.3321} \\ \hline \\ \hline

\end{tabular}%
}
\caption{\method\ vs. Baselines for Satirical Image Understanding Task on \textit{\textbf{YesBut}} Dataset and Meme Captioning Task on MemeCap Dataset in zero-shot setting. The best and second-best results for a VLM  are highlighted in \textbf{bold} and \textit{italic} respectively}
\label{tab:humor-und}
\end{table}

We observe that - (1) Given MiniCPM as the VLM, \method\ gives the best Avg. Score, boosting the Avg. Score by $0.72\%$ and $5.81\%$ compared to the best baseline on the Satire and Meme Understanding Tasks respectively\footnote{all improvements are significant according to independent two-sample t-test}, suggesting that \method\ is useful for improving humor comprehension of small open-source VLMs \noteng{I am not sure about this observation, please check}(2) \method\ gives a consistent improvement across all baselines when using GPT-4o-mini as the VLM (unlike MiniCPM), which could be attributed to superior code generation and understanding capabilities compared to MiniCPM (as \method\ uses Code-based reasoning) due to a larger context window length (GPT-4o-mini: 128k tokens, MiniCPM: 32k tokens), even though both models have similar sizes \cite{yao2024minicpm, abacha2024medec} (3) When using GPT-4o as the VLM, \method\ is the best compared to baselines across the tasks, and also is better than when using GPT-4o-mini and MiniCPM, which could be attributed to the larger model size of GPT-4o (4) According to Fig. \ref{fig:yesbut_impr}, \method\ gives the highest performance improvement compared to the best baseline on the \textit{\textbf{YesBut}} Dataset when using GPT-4o-mini VLM, reinforcing the notion that \method\ is effective even when the model is small. 

\pg{The BLEU scores are quite poor, this would need to be discussed. Also, can't you use LLM-as-a-judge?}

\subsubsection{Few-Shot In-Context Learning Setting}


Table \ref{tab:humor-und-few-shot} shows the performance of \method\ vs. several baselines on the Satirical Image Understanding and Meme captioning Tasks in few-shot setting across different number of in-context examples and VLMs. We observe that - (1) \method\ outperforms the baselines across all metrics in 2-shot setting, suggesting that a little in-context supervision is sufficient for effective causal reasoning graph generation across VLMs (2) \method\ gives the best Avg. Score when using the GPT-4o VLM, which is again due to its larger size. (3) When increasing the in-context examples, there is a diminishing return in performance improvement (e.g. \method\ gives a performance improvement of $11.66\%$ in 0-shot, $10.14\%$ in 2-shot, and $5.86\%$ in 5-shot setting compared to CoT when using GPT-4o VLM), suggesting that increasing in-context examples does not necessarily improve VLM's understanding capability. (4) As model size increases, having more in-context examples does not necessarily translate to better performance of \method, probably due to better parametric knowledge of the model (5) \method\ gives better performance improvement on Meme Captioning compared to Satire Understanding, suggesting that a supporting text (here, meme title) as an input in addition to the image in in-context examples leads to better supervision.    

\begin{table}[H]
\centering
\resizebox{\linewidth}{!}{%
\begin{tabular}{c|c|cccc|cccc}
\hline
\multirow{2}{*}{\textbf{VLM}} & \multirow{2}{*}{\textbf{Method}} & \multicolumn{4}{c|}{\textbf{Satirical Image Understanding}} & \multicolumn{4}{c}{\textbf{Meme Captioning}} \\ \cline{3-10}
                             &                                  & \textbf{ROUGE-L} & \textbf{BLEU} & \textbf{BERTScore} & \textbf{Avg.} & \textbf{ROUGE-L} & \textbf{BLEU} & \textbf{BERTScore} & \textbf{Avg.} \\ \hline 

\multirow{3}{*}{\begin{tabular}[c]{@{}c@{}}MiniCPM\\(2-shot)\end{tabular}} 
& Vanilla    & 0.2187 & 0.0251 & 0.8763 & 0.3734 & 0.0832 & 0.0078 & 0.8365 & 0.3092 \\
& CoT       & \textit{0.2198} & \textit{0.0278} & \textit{0.8778} & \textit{0.3751} & \textit{0.1078} & \textit{0.0099} & \textit{0.8463} & \textit{0.3213}  \\
& \method       & \textbf{0.2274} & \textbf{0.0305} & \textbf{0.8791} & \textbf{0.379} & \textbf{0.134} & \textbf{0.0136} & \textbf{0.8541} & \textbf{0.3339}  \\ \hline
\multirow{3}{*}{\begin{tabular}[c]{@{}c@{}}MiniCPM\\(5-shot)\end{tabular}} 
& Vanilla    & \textbf{0.2508} & \textit{0.0351} & \textbf{0.8861} & \textbf{0.3907} & 0.0839 & 0.0079 & 0.837 & 0.3096 \\
& CoT       & 0.2404 & 0.035 & 0.8848 & 0.3867 & \textit{0.1183} & \textit{0.0106} & \textit{0.8471} & \textit{0.3253}  \\
& \method       & \textit{0.2414} & \textbf{0.0368} & \textit{0.8851} & \textit{0.3878} & \textbf{0.1353} & \textbf{0.0138} & \textbf{0.8546} & \textbf{0.3346}  \\ \hline \\ \hline

\multirow{3}{*}{\begin{tabular}[c]{@{}c@{}}GPT-4o-mini\\(2-shot)\end{tabular}}
& Vanilla    & 0.1837 & 0.0148 & 0.8654 & 0.3546 & 0.0935 & 0.0047 & 0.8467 & 0.315 \\
& CoT       & \textit{0.2068} & \textit{0.0242} & \textit{0.8729} & \textit{0.368} & \textit{0.1092} & \textit{0.0068} & \textit{0.8497} & \textit{0.3219}  \\
& \method       & \textbf{0.2384} & \textbf{0.0325} & \textbf{0.8814} & \textbf{0.3841} & \textbf{0.1267} & \textbf{0.0071} & \textbf{0.8573} & \textbf{0.3304} \\ \hline
\multirow{3}{*}{\begin{tabular}[c]{@{}c@{}}GPT-4o-mini\\(5-shot)\end{tabular}} 
& Vanilla    & 0.218 & 0.0235 & \textit{0.8773} & 0.3729 & 0.0937 & 0.0049 & 0.8471 & 0.3152 \\
& CoT       & \textit{0.2289} & \textbf{0.0296} & \textbf{0.8822} & \textbf{0.3802} & \textit{0.1053} & \textit{0.0074} & \textit{0.8543} & \textit{0.3223}  \\
& \method       & \textbf{0.2396} & \textit{0.0294} & 0.8567 & \textit{0.3752} & \textbf{0.13} & \textbf{0.0076} & \textbf{0.8584} & \textbf{0.332} \\ \hline \\ \hline

\multirow{3}{*}{\begin{tabular}[c]{@{}c@{}}GPT-4o\\(2-shot)\end{tabular}} 
& Vanilla    & \textit{0.2146} & \textit{0.0244} & 0.8415 & \textit{0.3602} & 0.1153 & 0.0065 & 0.8531 & 0.325 \\
& CoT       & 0.1831 & 0.0141 & \textit{0.8681} & 0.3551 & \textit{0.126} & \textit{0.0083} & \textit{0.8535} & \textit{0.3293}  \\
& \method       & \textbf{0.2534} & \textbf{0.0339} & \textbf{0.886} & \textbf{0.3911} & \textbf{0.1479} & \textbf{0.0095} & \textbf{0.8636} & \textbf{0.3403} \\ \hline
\multirow{3}{*}{\begin{tabular}[c]{@{}c@{}}GPT-4o\\(5-shot)\end{tabular}} 
& Vanilla    & \textit{0.231} & \textit{0.0266} & 0.8417 & 0.3664 & 0.1188 & 0.0068 & 0.8543 & 0.3266 \\
& CoT       & 0.206 & 0.0198 & \textit{0.8797} & \textit{0.3685} & \textit{0.1315} & \textit{0.009} & \textit{0.8578} & \textit{0.3328}  \\
& \method       & \textbf{0.2513} & \textbf{0.0318} & \textbf{0.8872} & \textbf{0.3901} & \textbf{0.1624} & \textbf{0.0106} & \textbf{0.8654} & \textbf{0.3461} \\ \hline \\ \hline

\end{tabular}%
}
\caption{\method\ vs. Baselines for Satirical Image Understanding Task on \textit{\textbf{YesBut}} Dataset and Meme Captioning Task on MemeCap Dataset in few-shot setting. The best and second-best results for a VLM and a particular number of in-context examples are highlighted in \textbf{bold} and \textit{italic} respectively}
\label{tab:humor-und-few-shot}
\end{table}

\subsection{\method\ on Humor Detection}


\begin{table}[H]
\centering
\begin{minipage}{0.48\linewidth}
\normalsize	
\centering
\scriptsize
\begin{tabular}{|c|c|c|c|c|}
\hline
\textbf{Metric} & \textbf{Vanilla} & \textbf{CoT} & \textbf{\method} & \textbf{Impr.} \\
\hline
\multicolumn{5}{|c|}{\textbf{0-shot}} \\
\hline
Accuracy & 47.42\% & 48.07\% &  49.48\%  & 2.93\%\\
F1 Score & 61.05\% & 61.69\% &  62.20\% & 0.83\%\\
\hline
\multicolumn{5}{|c|}{\textbf{2-shot}} \\
\hline
Accuracy & 47.81\% & 48.32\% &  49.88\% & 3.23\% \\
F1 Score & 61.22\% & 61.73\% & 62.38\% & 1.05\% \\
\hline
\multicolumn{5}{|c|}{\textbf{6-shot}} \\
\hline
Accuracy & 47.85\% & 48.61\% & 49.91\% & 2.67\% \\
F1 Score & 61.27\% & 61.85\% & 62.41\% & 0.9\% \\
\hline
\end{tabular}
\caption{Comparison of Vanilla, CoT and \method\ across different number of in-context examples on Sarcasm Detection in MMSD 2.0 Dataset using GPT-4o, along with performance metric improvements compared to best baseline}
\label{tab:mmsd-results}
\end{minipage}
\hfill
\begin{minipage}{0.48\linewidth}
\centering
\scriptsize
\begin{tabular}{|c|c|c|c|c|}
\hline
\textbf{Metric} & \textbf{Vanilla} & \textbf{CoT} & \textbf{\method} & \textbf{Impr.} \\
\hline
\multicolumn{5}{|c|}{\textbf{0-shot}} \\
\hline
Accuracy & 42.60\% & 42.7\% & 43.18\%  & 1.12\%\\
F1 Score & 59.69\% & 59.75\% & 59.97\% & 0.37\%\\
\hline
\multicolumn{5}{|c|}{\textbf{2-shot}} \\
\hline
Accuracy & 44.05\% & 44.39\% & 44.63\% & 0.54\% \\
F1 Score & 60.52\% & 60.68\% &  61.01\% & 0.54\% \\
\hline
\multicolumn{5}{|c|}{\textbf{6-shot}} \\
\hline
Accuracy & 44.91\% & 45.38\% &  45.57\% & 1.05\% \\
F1 Score & 61.08\% & 61.32\% &  61.56\% & 0.39\% \\
\hline
\end{tabular}
\caption{Comparison of Vanilla, CoT, and \method\ across different number of in-context examples on Satire Detection in YesBut Dataset using GPT-4o, along with performance metric improvements compared to best baseline}
\label{tab:yesbut-det-results}
\end{minipage}
\end{table}

Tables \ref{tab:mmsd-results} and \ref{tab:yesbut-det-results} show the performance improvement compared to the best baseline when using \method\ on GPT-4o for Sarcasm Detection on the MMSD 2.0 Dataset and Satire Detection on the \textit{\textbf{YesBut}} Dataset respectively. Note that intuitively, Humor Detection is easier than Humor Understanding - hence, we compare \method\ with Vanilla and the well-performing baseline of CoT for experiments on Humor Detection. We observe that \method\ performs consistently better across varying number of in-context examples. In addition, the performance improvement in detecting sarcasm is better than satire, which might be due to the additional supporting text in the sarcasm detection task providing additional supervision while generating a causal reasoning graph. 


\subsection{Dissecting \method's Reasoning Component}
To investigate the underlying factors contributing to the superior performance of \method, we specifically analyze the reasoning component ($\mathcal{R}$) generated by \method\  in comparison to that produced by baseline methods. Our analysis focuses on two key aspects:
(A) Whether the reasoning produced by \method\ contains richer and more novel information relative to the baselines, and
(B) Whether the final ground truth can be more effectively inferred from the reasoning provided by \method\ than from that of the baselines.

To assess (A), we employ pairwise dissimilarity/divergence based measures across \method\ and baselines - \textbf{KL Divergence between the token distributions} for measuring the extra amount of lexical information in $\mathcal{R}$ of one method compared to another, and a \textbf{sentence similarity-based dis-similarity score} for measuring the amount of semantically newer information in a method's $\mathcal{R}$ relative to another. Similarly, to assess (B), we employ an \textbf{LLM-as-a-judge \cite{zheng2023judging} approach to infer whether $\mathcal{Y}$ can be logically inferred from $\mathcal{R}$}. To ensure uniform evaluation, we apply \method\ and the baselines of CoT, CoD, and CCoT using MiniCPM in zero-shot setting to the Satirical Image Understanding Task on the \textit{\textbf{YesBut}} Dataset. These are described in detail as follows -

\subsubsection{KL Divergence between token distributions}

Given the texts $T_1, T_2$, 
$KL(T_1 ||T_2)$ and $KL(T_2 ||T_1)$
are calculated after tokenizing $T_1$ and $T_2$ by splitting with whitespace and making the tokens lowercase, and the set of all tokens so obtained from $T_1$ and $T_2$ forms the vocabulary (check \textit{\textbf{Algorithm \ref{alg:kl_divergence} in Section \ref{appendix:further-analysis} of Appendix}} for details). 
%
Since KL-divergence measures how well one distribution approximates another, a higher $KL(T_1 ||T_2)$ than $KL(T_2 ||T_1)$ suggests that the lexical distribution in 
$T_1$ deviates more from 
$T_2$ 's distribution than vice versa, possibly indicating that 
$T_1$  contains more varied or less predictable lexical content. \pg{Not sure. Any reference, or analysis? Even if it is true, why is it a good thing? The new content may be wrong.}
Table \ref{tab:kl-div} shows the KL-Divergence between reasoning components of \method\ and baselines. We see that \method\ has a greater or equal amount of newer lexical information compared to the 3 baselines.


\begin{table}[H]
  \centering
  \begin{minipage}[t]{0.48\textwidth}
    \centering
    \scriptsize
    \begin{tabular}{|c|c|c|}
      \hline
      $T_1$ & $T_2$ & $KL(T1||T2)$ \\
      \hline
      $\mathcal{R}_{\method}$ & $\mathcal{R}_{CoT}$  & \textbf{0.21} \\
      $\mathcal{R}_{CoT}$      & $\mathcal{R}_{\method}$ & 0.19 \\
      \hline
      $\mathcal{R}_{\method}$ & $\mathcal{R}_{CoD}$  & \textbf{0.21} \\
      $\mathcal{R}_{CoD}$      & $\mathcal{R}_{\method}$ & 0.20 \\
      \hline
      $\mathcal{R}_{\method}$ & $\mathcal{R}_{CCoT}$ & \textbf{0.25} \\
      $\mathcal{R}_{CCoT}$     & $\mathcal{R}_{\method}$ & \textbf{0.25} \\
      \hline
    \end{tabular}
    \caption{KL-Divergence between methods (Method name is in subscript)}
    \label{tab:kl-div}
  \end{minipage}\hfill
  \begin{minipage}[t]{0.48\textwidth}
    \centering
    \scriptsize
    \begin{tabular}{|c|c|c|}
      \hline
      $T_1$ & $T_2$ & $LSF(T_1||T_2)$ \\
      \hline
      $\mathcal{R}_{\method}$ & $\mathcal{R}_{CoT}$  & \textbf{1} \\
      $\mathcal{R}_{CoT}$      & $\mathcal{R}_{\method}$ & 0.85 \\
      \hline
      $\mathcal{R}_{\method}$ & $\mathcal{R}_{CoD}$  & \textbf{1} \\
      $\mathcal{R}_{CoD}$      & $\mathcal{R}_{\method}$ & 0.84 \\
      \hline
      $\mathcal{R}_{\method}$ & $\mathcal{R}_{CCoT}$ & \textbf{0.98} \\
      $\mathcal{R}_{CCoT}$     & $\mathcal{R}_{\method}$ & 0.94 \\
      \hline
    \end{tabular}
    \caption{LSF between methods (Method name is in subscript)}
    \label{tab:lsf}
  \end{minipage}
\end{table}




\subsubsection{Sentence similarity-based dissimilarity score}
In sentence-based similarity, denoted as $LSF(T_y \,\|\, T_x)$, we compute the fraction of sentences in $T_x$ that are semantically similar to at least one sentence in $T_y$. This metric reflects the extent to which the content of $T_x$ is covered by $T_y$. Sentence similarity is evaluated using Sentence-BERT \cite{sbert}, where two sentences are considered similar if the cosine similarity between their respective embeddings exceeds 0.5 (refer to \textit{\textbf{Algorithm~\ref{alg:low_similarity_fraction} in Section~\ref{appendix:further-analysis} of the Appendix}} for details).
Similar to KL-Divergence, if $LSF(T_1 ||T_2) > LSF(T_2 ||T_1)$, the additional semantic information in $T_1$ is greater than that in $T_2$. Table \ref{tab:lsf} shows LSF between reasoning components of \method\ and baselines. We can see that \method\ has a greater amount of newer semantic information compared to the 3 baselines.

\pg{But shouldn't we also compute these with respect to the ground truth for each of the method?}

\subsubsection{LLM-as-a-judge approach to infer whether $\mathcal{Y}$ can be logically inferred from $\mathcal{R}$}

\begin{wraptable}{r}{0.5\textwidth}
  \centering
  \footnotesize
  \caption{\textsc{InferScore}: \method\ vs. Baselines}
  \label{tab:inferscores}
  \begin{tabular}{@{}l|c@{}}
    \toprule
    Method & \textsc{InferScore} \\
    \midrule
    CoT    & 40.78 \\
    CoD    & 40.68 \\
    CCoT   & 37.64 \\
    \method & \textbf{45.11} \\
    \bottomrule
  \end{tabular}
\end{wraptable}

A detailed prompt containing $\mathcal{R}$ and $\mathcal{Y}$ (as well as details of the Humor Comprehension task) is given as input to GPT-4 \cite{gpt4}, and a prediction of $1$ means that $\mathcal{Y}$ can be logically inferred from $\mathcal{R}$, and $0$ means otherwise (here, we rely on the parametric knowledge of GPT-4). Given a method, this is done for all dataset samples and the percentage of samples giving a prediction of $1$ is a score (referred to here as \textsc{InferScore}) representing the effectiveness of the generated reasoning component of the method. Table \ref{tab:inferscores} shows that \method\ attains the highest \textsc{InferScore}.

\section{Conclusion}

This work shows the effectiveness of using VLMs to generate Causal Reasoning Graphs, thereby enhancing Multimodal Humor Comprehension. Our approach \method\ leverages causal reasoning graphs to systematically model the relationships among events, entities, and contextual cues in multimodal inputs. Experiments validate the superiority of \method\ across VLMs in comparison to other reasoning-based baselines such as CoT, CoD, and CCoT in both zero-shot and few-shot in-context learning settings. Notably, Causal Reasoning Graphs enhance the amount of relevant reasoning information that is required to better understand and detect humor in multimodal scenarios.

\bibliography{custom}

\appendix

\section*{Appendix}

\section{Rectified vs. GPT-4o generated Causal Reasoning Graph}
\label{sec:manual-rectify}

\begin{figure}[H]
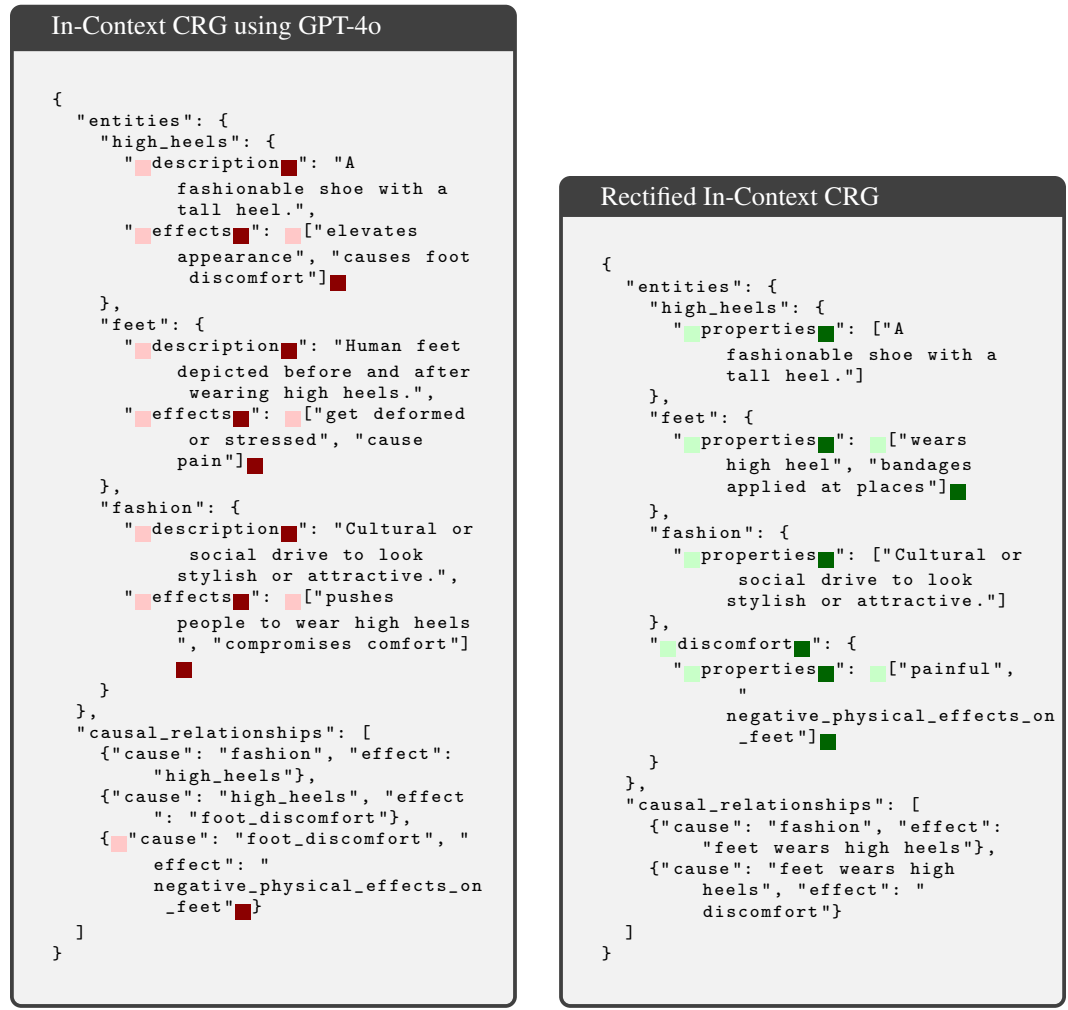

\centering

\begin{subfigure}{0.48\textwidth}
  \begin{tcolorbox}[title=In-Context CRG using GPT-4o]
\begin{lstlisting}[style=jsonstyle]
{
  "entities": {
    "high_heels": {
      "@R@description@S@": "A fashionable shoe with a tall heel.",
      "@R@effects@S@": @R@["elevates appearance", "causes foot discomfort"]@S@
    },
    "feet": {
      "@R@description@S@": "Human feet depicted before and after wearing high heels.",
      "@R@effects@S@": @R@["get deformed or stressed", "cause pain"]@S@
    },
    "fashion": {
      "@R@description@S@": "Cultural or social drive to look stylish or attractive.",
      "@R@effects@S@": @R@["pushes people to wear high heels", "compromises comfort"]@S@
    }
  },
  "causal_relationships": [
    {"cause": "fashion", "effect": "high_heels"},
    {"cause": "high_heels", "effect": "foot_discomfort"},
    {@R@"cause": "foot_discomfort", "effect": "negative_physical_effects_on _feet"@S@}
  ]
}
\end{lstlisting}
  \end{tcolorbox}
\end{subfigure}
\hfill
\begin{subfigure}{0.48\textwidth}
  \begin{tcolorbox}[title=Rectified In-Context CRG]
\begin{lstlisting}[style=jsonstyle]
{
  "entities": {
    "high_heels": {
      "@G@properties@H@": ["A fashionable shoe with a tall heel."]
    },
    "feet": {
      "@G@properties@H@": @G@["wears high heel", "bandages applied at places"]@H@
    },
    "fashion": {
      "@G@properties@H@": ["Cultural or social drive to look stylish or attractive."]
    },
    "@G@discomfort@H@": {
      "@G@properties@H@": @G@["painful", "negative_physical_effects_on _feet"]@H@
    }
  },
  "causal_relationships": [
    {"cause": "fashion", "effect": "feet wears high heels"},
    {"cause": "feet wears high heels", "effect": "discomfort"}
  ]
}
\end{lstlisting}
  \end{tcolorbox}
\end{subfigure}
\caption{Comparison of In-Context CRGs (Causal Reasoning Graphs): GPT-4o generated vs. Rectified version. The text spans with ``\colorbox{hlred}{}'' on the left and ``\colorbox{darkred}{}'' on the right in the GPT-4o generated CRG are rectified (removed/modified) to obtain the rectified CRG, where the spans with ``\colorbox{hlgreen}{}'' on the left and ``\colorbox{darkgreen}{}'' on the right are the resulting changes}
\label{fig:rectify-crg}
\end{figure}

Fig. \ref{fig:rectify-crg} compares in-context CRGs before and after rectification, highlighting the changes that took place.

\section{Example of Humor Understanding and Humor Detection Tasks}
\label{sec:humor-example}

\begin{figure}[H]
    \centering
    \includegraphics[width=\linewidth]{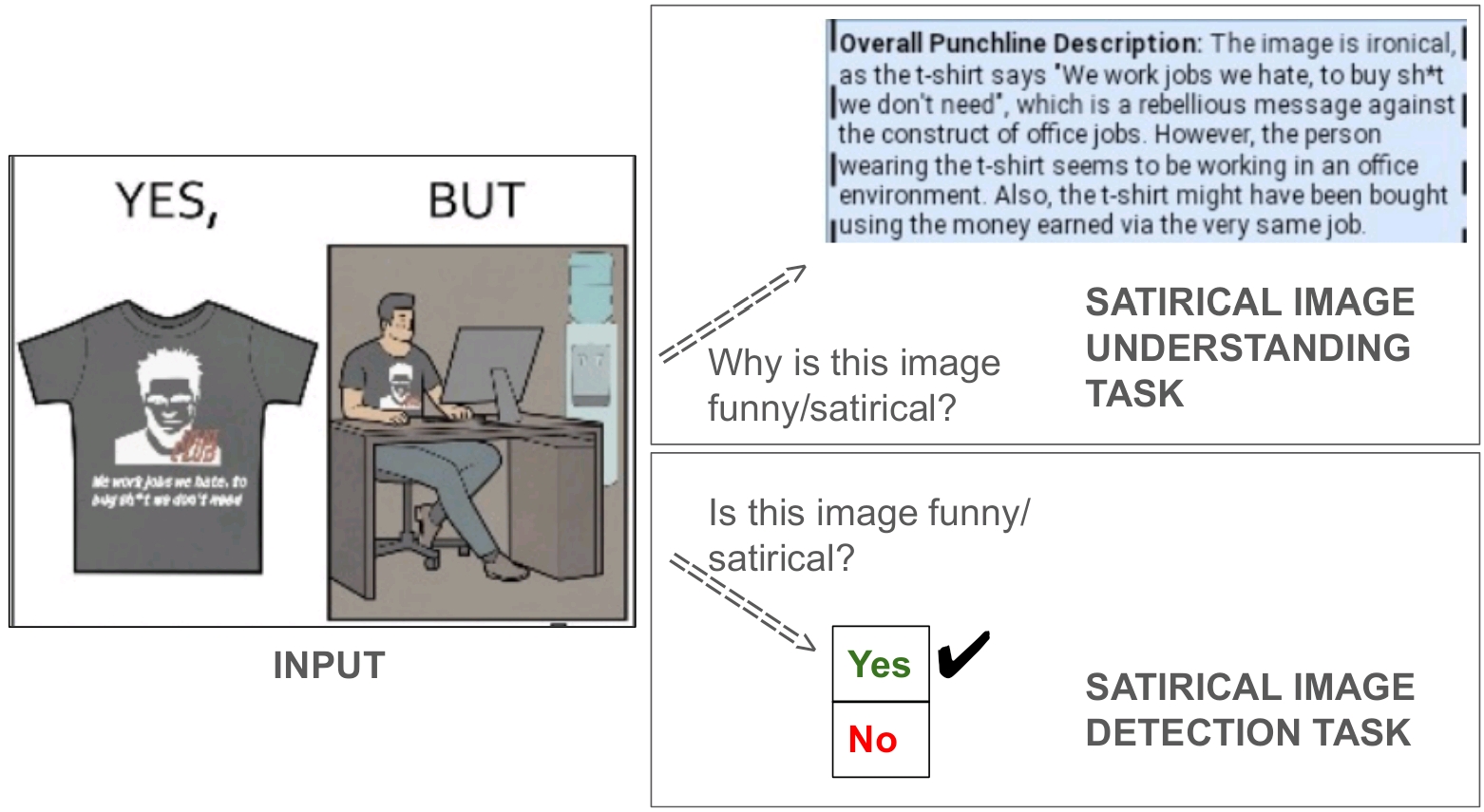}
    \caption{Example for Humor (in this case, Satire) Understanding and Detection Tasks}
    \label{fig:humor-example}
\end{figure}

Fig. \ref{fig:humor-example} shows examples for Humor (in this case, Satire) Understanding and Detection Tasks, listing the input, task, and ground truth.

\section{Task-specific Queries}
\label{sec:task-spec-q}

\noindent \textbf{Satirical Image Understanding. }\texttt{Why is this image funny/satirical?}

\noindent \textbf{Meme Caption Generation. }\texttt{Why is this meme funny?}

\noindent \textbf{Satirical Image Detection. }\texttt{You are an AI expert in detecting humor
or satire. User gives you an image, and you have to make a choice "Y" or "N". Instructions: Users image has 2 halves called yes and but, and the combination of those might make no sense at all, or be extremely funny. Your job is to find out which one it is and output Y if its EXTREMELY funny and N for otherwise. Output format: one character, exactly either "Y" or "N"}

\noindent \textbf{Multimodal Sarcasm Detection. }\texttt{Is this image with the text funny/sarcastic? Give your final answer as `<<YES>>' or `<<NO>>'}

\section{Example of an in-context example for CoT}
\label{appendix:cot}

\begin{figure}[H]
    \centering
    \includegraphics[width=0.5\linewidth]{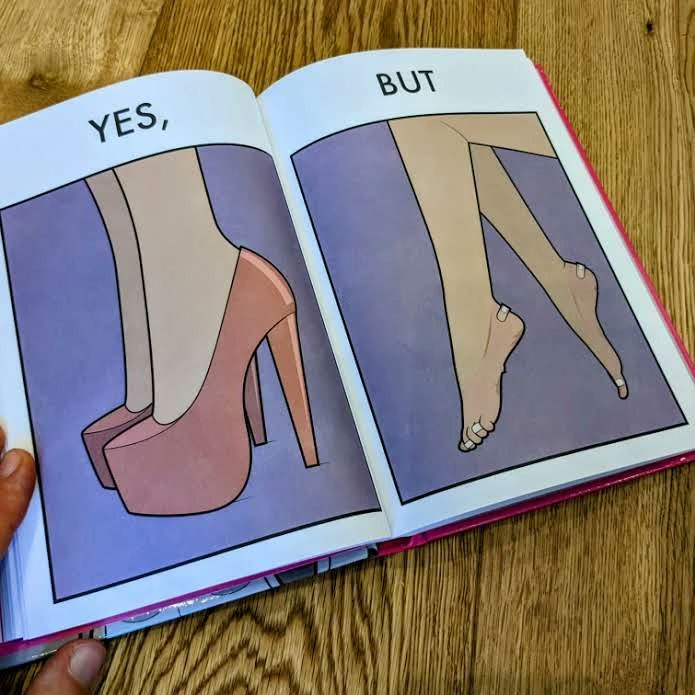}
    \caption{Example Image from the YesBut Dataset}
    \label{fig:cot-ice-image}
\end{figure}

Corresponding to Fig. \ref{fig:cot-ice-image}, the rationale and ground truth for CoT can be described as follows -

\noindent \textbf{Rationale. }

\texttt{Subject: Fashion footwear choices
      Premise: The image contrasts how high heels appear aesthetically versus their physical impact}
      
\texttt{Punchline: What looks beautiful and fashionable in design causes visible physical discomfort and strain}

\texttt{Irony: People willingly sacrifice comfort for style and appearance}

\texttt{Satirical commentary: The stark contrast between the elegant presentation of fashionable shoes and the resulting physical consequences they inflict on the wearer's feet}

\noindent \textbf{Ground Truth Answer. }The image is funny since it shows how wearing high heels in the name of fashion ends up causing a lot of physical discomfort to the user.    

\section{Ablation Analysis}
\label{sec:ablation}

\begin{table}[H]
\centering
\resizebox{0.8\textwidth}{!}{
\begin{tabular}{c|c|cccc}
\hline
\textbf{No.\ of ICE} & \textbf{Ablation} & \textbf{ROUGE-L} & \textbf{BLEU} & \textbf{BERTScore} & \textbf{Avg. Score} \\ \hline

\multirow{2}{*}{0} 
  & WITH DEFN. & 0.2131 & 0.0211 &    \textbf{0.8718} & 0.3687  \\ 
  & \method   & \textbf{0.2219} & \textbf{0.0245} &   0.8715 & \textbf{0.3726}  \\ \hline

\multirow{3}{*}{2} 
  & WITH DEFN.   & 0.2406 & 0.0309 &   \textit{0.8816} & \textit{0.3844} \\ 
  & UNRECTIFIED  & \textit{0.2492} & \textit{0.0327} &  0.8495 & 0.3771  \\ 
  & \method      & \textbf{0.2534} & \textbf{0.0339} &  \textbf{0.886} &  \textbf{0.3911} \\ \hline

\multirow{3}{*}{5} 
  & WITH DEFN.   & \textit{0.2478} & 0.0308 &   \textit{0.8844} & \textit{0.3877} \\ 
  & UNRECTIFIED  & 0.2466 & \textit{0.0310} &   0.8497  & 0.3758 \\ 
  & \method      & \textbf{0.2513} & \textbf{0.0318} &   \textbf{0.8872} & \textbf{0.3901} \\ \hline

\end{tabular}
}
\caption{Ablation Analysis of \method\ on GPT-4o. The best and second-best results for a particular number of in-context examples are highlighted in \textbf{bold} and \textit{italic} respectively (ICE - In-Context Examples)}
\label{tab:ablation}
\end{table}

We compare \method\ with - (a) \textbf{WITH DEFN. }A detailed definition of the causal reasoning graph (mentioned \textit{\textbf{in Section \ref{sec:crg-defn} of Appendix}}) is added to the prompt of \method\ (b) \textbf{UNRECTIFIED \method. }Causal Reasoning Graphs of in-context examples are not rectified manually in \method\ in few-shot setting 
Note that \textbf{WITH DEFN.} is carried out in zero-shot setting; all the ablations are carried out in in-context learning setting otherwise. The ablation analysis is carried out on GPT-4o in 0, 2, and 5-shot settings. We can infer from Table \ref{tab:ablation} that - (1) \method\ shows a consistent improvement on the lexical ROUGE-L, BLEU metrics, as well as the Avg. Score (2) In in-context learning setting, \method\ performs better than UNRECTIFIED across all metrics, showing the importance of manually rectifying the causal reasoning graphs in in-context examples (3) \method\ performs better than WITH DEFN. on 3/4 and 4/4 metrics for zero-shot and in-context learning settings respectively, suggesting that - using the definition of causal reasoning graph along with the task-specific query might be confusing for the VLM.

\subsection{Definition of Causal Reasoning Graph used in WITH DEFN.}
\label{sec:crg-defn}

\texttt{1. Entities: There is a set of entities (listed in "entities") — each entity can have properties that are either descriptions of/adjectives/adverbs qualifying that entity or (non-causal) relations with other entities. These are listed under “properties” attribute of each entity.
e.g entity: ANIMAL, "properties": ["Cats and Dogs", “pet of HUMAN”, “hides under FURNITURE”], note that HUMAN AND FURNITURE are other entities
entity: FIREWORKS, "properties": [ "Bright and colorful explosions in sky", "burnt by HUMAN”] 
note that the (non—causal) relationships are bidirectional, e.g. FIRECRACKERS (burnt by) HUMAN, HUMAN (burns) FIRECRACKER are same relationships. This relationship is present in "properties" list of any ONE of these entities e.g ("burns FIRECRACKERS" belongs to HUMAN[“properties"]) OR ("burnt by HUM" belongs to FIRECRACKER[“properties"]), BUT NOT BOTH} 

\texttt{2. CAUSAL relationships: listed under ``causal\_relationships''. First, we define an EVENT. A collection of entities (along with their (non—causal) relationships) describes an EVENT which is typically of the form "X (optionally) does Y (optionally) with/for/to Z", (a single entity can also be an EVENT) — an EVENT is basically a macro node and a causal relation is defined between events.
A causal relation is listed under "causal\_relationships" as a dictionary {"cause": EVENT\_1, "effect': EVENT\_2}. Each event is expressed in natural language which tells what the collection of entities means, for instance, ``X (optionally) does Y (optionally) with/for/to Z''. For example, {"cause": “HUMAN burns FIRECRACKER S”, "effect': “ANIMALS” are frightened}}

\section{Further Analysis of \method's Reasoning Component}
\label{appendix:further-analysis}

\begin{minipage}[t]{0.48\textwidth}
\begin{algorithm}[H]
\caption{KL-Divergence between token distributions}
\label{alg:kl_divergence}
\KwIn{Texts $T_1, T_2$; smoothing parameter $\alpha > 0$}
\KwOut{$KL(T_1 || T_2)$}

\BlankLine
\SetKwProg{Fn}{Function}{}{}
\Fn{\textsc{PreprocessAndTokenize}(Text)}{
  \Return{lowercase(Text) split on whitespace}
}
\BlankLine

\tcp{Tokenize both texts}
$\mathcal{T}_1 \leftarrow$ \textsc{PreprocessAndTokenize}($T_1$)\;
$\mathcal{T}_2 \leftarrow$ \textsc{PreprocessAndTokenize}($T_2$)\;

\tcp{Build vocabulary}
$\mathcal{V} \leftarrow \mathcal{T}_1 \cup \mathcal{T}_2$\; 
$V \leftarrow |\mathcal{V}|$\;

\tcp{Count tokens}
\ForEach{$i \in \{1,2\}$}{
  $\forall w \in \mathcal{V}:\; c_i(w) \leftarrow \#\{t\in\mathcal{T}_i : t = w\}$\;
}

\tcp{Compute smoothed distributions}
\For{$i\in\{1,2\}$}{
  $\displaystyle Z_i \leftarrow |\mathcal{T}_i| + \alpha\,V$\;
  $\displaystyle \forall w\in\mathcal{V}:\;
    \mathcal{D}_i(w) \leftarrow \frac{c_i(w) + \alpha}{Z_i}$\;
}
$\mathcal{P} \leftarrow \mathcal{D}_1$, $\mathcal{Q} \leftarrow \mathcal{D}_2$\;

\tcp{Compute KL divergence}
$\displaystyle D \leftarrow 0$\;
\ForEach{$w\in\mathcal{V}$}{
  $D \leftarrow D \;+\; \mathcal{P}(w)\,\log\frac{\mathcal{P}(w)}{\mathcal{Q}(w)}$\;
}

\Return{$D$}
\end{algorithm}
\end{minipage}
\hfill
\begin{minipage}[t]{0.48\textwidth}
\begin{algorithm}[H]
\caption{Low Similarity Fraction (LSF)}
\label{alg:low_similarity_fraction}
\KwIn{Texts $T_1, T_2$; Upper Bound $U \in [0,1]$}
\KwOut{$LSF(T_1||T_2)$: fraction of sentences in $T_1$ whose average similarity to $T_2$ is below $U$}

\BlankLine
\SetKwProg{Fn}{Function}{}{}
\Fn{\textsc{PreprocessSentences}(\,$T$)}{
  \Return{$\mathrm{nltk.sent\_tokenize}(T)$}
}
\BlankLine
\Fn{\textsc{EmbedSentences}(\,$S$)}{
  \Return{$[\,\mathrm{SentenceBert}(s)\;\forall\;s\in S\,]$}
}
\BlankLine

\tcp{Sentence‐level preprocessing}
$S_1 \leftarrow \textsc{PreprocessSentences}(T_1)$\;
$S_2 \leftarrow \textsc{PreprocessSentences}(T_2)$\;
$m \leftarrow |S_1|,\; n \leftarrow |S_2|$\;

\tcp{Compute sentence embeddings}
$E_1 \leftarrow \textsc{EmbedSentences}(S_1)$\;
$E_2 \leftarrow \textsc{EmbedSentences}(S_2)$\;

\tcp{Count low‐similarity sentences}
$count \leftarrow 0$\;
\ForEach{$e_1\in E_1$}{
  \tcp{Cosine similarities to all sentences in $T_2$}
  $\mathrm{simList} \leftarrow [\,\mathrm{cosineSim}(e_1, e_2)\;\forall\;e_2\in E_2]$\;
  $\bar s \leftarrow \frac{1}{n}\sum \mathrm{simList}$\;
  \If{$\bar s < U$}{
    $count \leftarrow count + 1$\;
  }
}

\tcp{Compute and return the fraction}
\Return{$\tfrac{count}{m}$}
\end{algorithm}
\end{minipage}

\end{document}